\documentclass[lettersize,journal]{IEEEtran}
\usepackage{amsmath,amsfonts}
\usepackage{algorithmic}
\usepackage{algorithm}
\usepackage{array}
\usepackage[caption=false,font=normalsize,labelfont=sf,textfont=sf]{subfig}
\usepackage{textcomp}
\usepackage{stfloats}
\usepackage{url}
\usepackage{verbatim}
\usepackage{graphicx}
\usepackage{cite}
\usepackage{xcolor}

\begin{document}

\title{Beyond Class Tokens: LLM-guided Dominant Property Mining for Few-shot Classification}

\author{Wei Zhuo, Runjie Luo, Wufeng Xue,~\IEEEmembership{Member, IEEE}, Linlin Shen\textsuperscript{\dag},~\IEEEmembership{Senior Member, IEEE}
        % <-this % stops a space
\thanks{\textsuperscript{\dag} Linlin Shen is the corresponding author.}
\thanks{Wei Zhuo, Runjie Luo, Linlin Shen are with the School of Artificial Intelligence and the National Engineering Laboratory of Big Data System Computing Technology, Shenzhen University, Shenzhen 518060, China (e-mail: weizhuo@szu.edu.cn, 
%alterorange175@gmail.com, 
2310533030@email.szu.edu.cn,
llshen@szu.edu.cn). Wufeng Xue is with School of Biomedical Engineering, Shenzhen University Medical School, Shenzhen University, Shenzhen 518060,
China (e-mail: xuewf@szu.edu.cn).}
}

\markboth{Journal of \LaTeX\ Class Files,~Vol.~XX, No.~XX, XXXX~2025}%
{Shell \MakeLowercase{\textit{et al.}}: A Sample Article Using IEEEtran.cls for IEEE Journals}


\maketitle

\section{abstract}
Few-shot Learning (FSL), which endeavors to develop the generalization ability for recognizing novel classes using only a few images, faces significant challenges due to data scarcity. Recent CLIP-like methods based on contrastive language-image pertaining mitigate the issue by leveraging textual representation of the class name for unseen image discovery. Despite the achieved success, simply aligning visual representations to class name embeddings would compromise the visual diversity for novel class discrimination. To this end, we proposed a novel Few-Shot Learning (FSL) method (BCT-CLIP) that explores \textbf{dominating properties} via contrastive learning beyond simply using class tokens. Through leveraging LLM-based prior knowledge, our method pushes forward FSL with comprehensive structural image representations, including both global category representation and the patch-aware property embeddings.  In particular, we presented a novel multi-property generator (MPG) with patch-aware cross-attentions to generate multiple visual property tokens, a Large-Language Model (LLM)-assistant retrieval procedure with clustering-based pruning to obtain dominating property descriptions, and a new contrastive learning strategy for property-token learning.  The superior performances on the 11 widely used datasets demonstrate that our investigation of dominating properties advances discriminative class-specific representation learning and few-shot classification. \\

\begin{IEEEkeywords}
Few-shot Learning, Visual-Language Pretraining, LLM-based Property Learning
\end{IEEEkeywords}

\section{Introduction}
Few-shot learning (FSL)~\cite{koch2015siamese,matching_net,protonet}, due to its efficient practical manner, has gained increasing attention and laid the foundation for subsequent vision tasks such as detection and segmentation
~\cite{u_net,meta_rcnn,fsod,latent-mining}. 
Instead of leveraging extensive labeled data in supervised learning~\cite{alexnet, vgg, ResNet}, few-shot learning (FSL) aims to develop generalized classification models that can be adapted rapidly to recognize unseen classes of novel tasks by providing just a few labeled images. 
However, the scarcity of novel data severely limits the richness of representations and the reliability of class discrimination on novel tasks. In view of this, a generalized powerful pretrained model is indispensable. To this end, recent works have extensively explored leveraging the Contrastive Image-Language Pretraining (CLIP) for FSL.

\begin{figure}[tp]
    \centering
    \includegraphics[width=\linewidth]{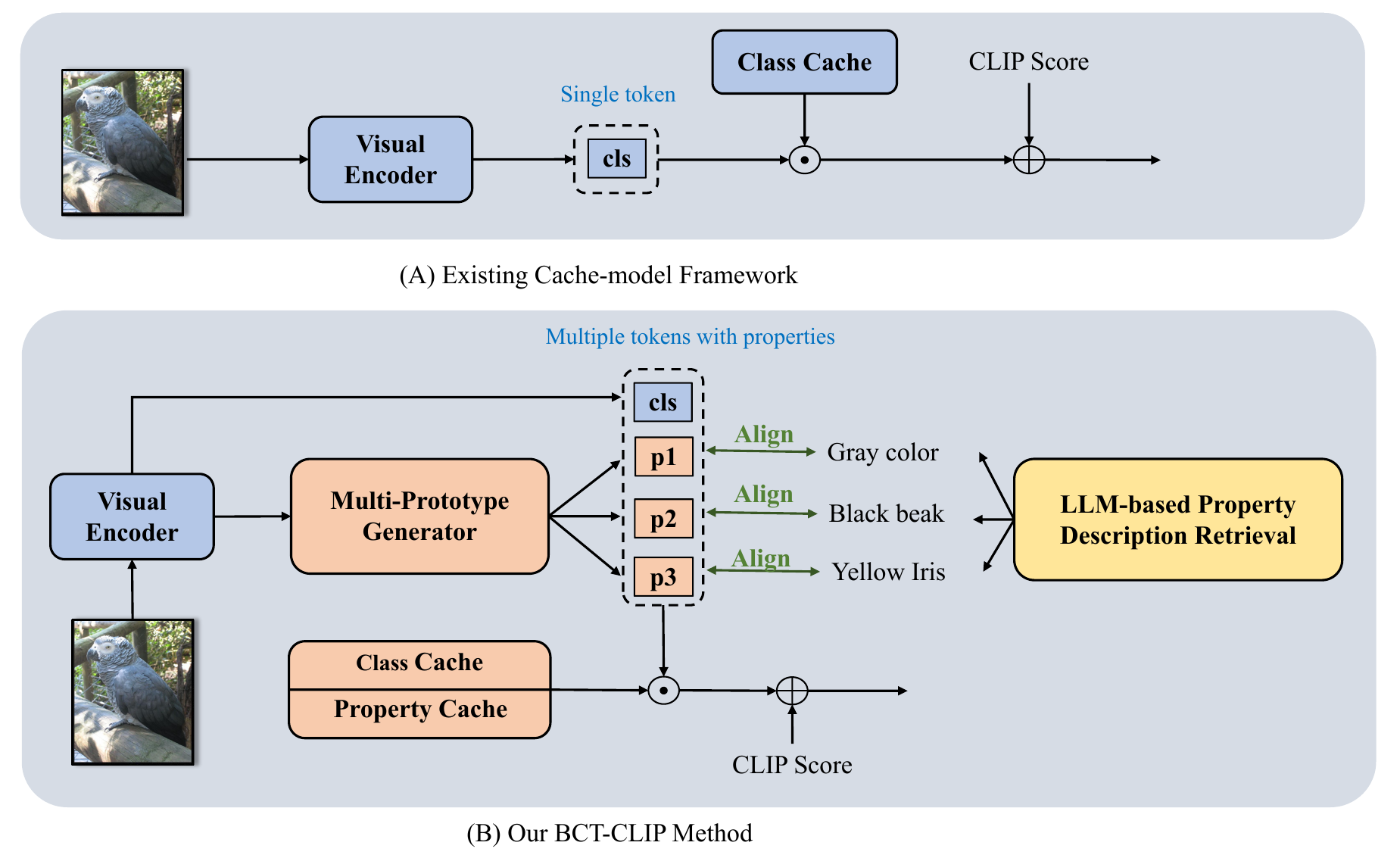}
    \caption{ 
    \textbf{Comparison with existing CLIP-based FSL methods. }
    Instead of using one single prototype for novel class, our approach introduces a multi-property framework, significantly enriching the representation of novel classes. This is enabled by our efficient Multi-Property Generator (MPG) and contrastive learning, which leverages properties retrieved from Large Language Models (LLMs). 
    %further boosting performance when combined with CLIPZero scores.
    }
    \label{intro_fig}
    \vspace{-16pt}
\end{figure}

CLIP is a multimodal foundation model whose image and text encoders are jointly trained to align visual and language representations on large-scale image–text pairs. Recent few-shot learning methods leverage this cross-modal alignment to match query images with textual prompts, typically class names. Generally, existing CLIP-based FSL can be categorized into two groups: prompt-based methods and the cache-model methods. 
Based on CLIP, early prompt-based works~\cite{CoOp,zhou2022conditional} improve image recognition by introducing trainable textual prompting vectors or an additional adapter fine-tuned on a few reference (support) images. Despite its effectiveness, a single textual embedding of class names cannot fully represent category details, significantly limiting models' discriminative power for fine-grained categories, which often have similar but varying details. 
To this end, studies like~\cite{CuLP,CuLP_Hierarchical} gather more detailed descriptions of object attributes and context by asking LLMs. The extended descriptions are generally used as prompts for visual-language matching. 
These methods, however, typically overlook pruning generic descriptions and noisy responses, resulting in compromised discrimination ability. In addition, since CLIP is trained on image-level representations instead of dense embeddings, its visual encoders failed to align with fine-grained descriptions, which is a core reason for the suboptimal performance on the extended prompts. We dub this phenomenon as \textbf{\textit{fine-grained alignment deficiency}}. This is because, despite obtaining fine-grained text prompts, the class token of the query image lacks sufficient ability to match with them. 

For these methods, embedding of the textual prompt practically serves as the classifier applied to the query image features. Rather than relying on a simple linear classifier as done in prompt-based methods, cache-modal methods, such as~\cite{Tip-adapter, tipx, PMPro}, develop an enhanced classification model, i.e., a key-value cache model, which is a collection of the visual embeddings of all few-shot support images and their one-hot labels. This model is further enhanced by training.  Its nature is to learn multiple classifiers per category for integrated decision. \textbf{\textit{These prompt-based and cache-model methods, however, primarily focus on improving classifiers through enhanced prompts or stronger cache models, while neglecting the enhancement of query image representation, typically relying solely on the image's class tokens for classification.} }

Unlike these methods, our work aims to enhance image representations by incorporating additional property-level representations. We argue that the class token potentially overlooks crucial spatial information for fine-grained discrimination, while learning representation for class properties, which are typically manifested through localized attributes such as shape, color, or texture, can compensate for this shortcoming. 
Generating property representations, however, is non-trivial, which faces two key challenges: (1) the absence of an effective network for property-specific feature generation; and (2) the lack of ground-truth property descriptions for supervised objectives.

To address these challenges, we introduce a novel approach, termed \textbf{BCT-CLIP}, that introduces LLM-guided dominant property mining \textbf{B}eyond merely using \textbf{C}lass \textbf{T}okens based on \textbf{CLIP}. 
This approach introduced a novel Multi-Property Generator (MPG) to generate property tokens and an LLM-based property-description retrieval procedure to achieve annotations for supervision. A new contrastive learning strategy is also introduced to match property tokens and its corresponding descriptions. By incorporating additional visual property tokens, our method efficiently enhances the query representations and mitigates the \textit{fine-grained alignment deficiency}.

Specifically, motivated by the observation that properties are often reflected in localized details, our Multi-Property Generator (MPG) is designed to embed a set of learnable vectors (property tokens) with patch features through group-wised cross-attention blocks. 
To achieve class property descriptions for supervision, we design an LLM-based pipeline involving clustering and support-based filtering to prune noisy descriptions and select dominating class-specific properties. Note that, compared to previous prompt-based methods~\cite{CuLP,CuLP_Hierarchical}, our method employs an efficient pruning strategy, ensuring that the selected property descriptions accurately reflect the class's true characteristics. Given these data, a new contrastive learning-based strategy is proposed to pull close the property tokens with their positive descriptions and push them away from the negative ones, in the end enabling fine-grained visual-text alignment for discrimination. A hybrid cache model is further learned to output class scores. By combining property tokens with the class token, our method learns comprehensive structural-semantic representations for precise classification.

Overall, our contribution lies in four folds:
\begin{itemize}
\item 
This approach introduced a novel Few-Shot Learning (FSL) architecture, BCT-CLIP, that advances FSL with advantage of LLM-prior knowledge to enable comprehensive structural semantical representations, including both global category representation and the localized property embeddings per image.
\item We propose a novel Multi-Property Generator (MPG), an LLM-based property description retrieval procedure with clustering and pruning,  and an effective contrastive learning strategy for property token learning. 
\item We introduce a hybrid cache model that incorporates both class and property embeddings for classification. 
\item Extensive experiments of few-shot classification are conducted on ImageNet and 10 more datasets, demonstrating the superiority of our enriched representations and the robust capability in domain generalization.
\end{itemize}

In the following sections, we present the related work in Section III, and a preliminary about cache model in Section IV. Subsequently, we introduce our method in detail. In particular, we first present the Multi-Property Generator (MPG) in Section V-B, the LLM-based pipeline for property annotation in Section V-C, and then the training objectives for contrastive learning in Section V-D. Section V-E presents our classification model, i.e., the hybrid cache model. The aforementioned sections illustrate the complete training pipeline. The scoring mechanism for inference is introduced in Section V-F.  

\section{Related Work}
\subsection{Classical Few-Shot Learning} 
\label{sec-classical-FSL}

Few-shot learning (FSL)~\cite{munkhdalai2017meta,protonet} aims to learn discriminative abilities for unseen classes given limited samples. To learn general model for FSL~\cite{rombach2022high}, meta-learning and transfer learning have been widely studied. 
In the FSL field, Meta-learning is proposed to learn transferable priors via episodic prototypical training~\cite{Reptile,protonet,matching_net,koch2015siamese,zhu2020attribute} on extensive support-query pairs. In particular, early work~\cite{koch2015siamese} constructs a dual-branch Siamese network to extract support prototype vectors and the query features, individually, for matching. This architecture is followed by many later works. 
However, meta-learning is usually limited by the intricate episodic-training mechanism and substantial computational cost. Unlike meta-learning, transfer learning employs simple conventional training, which pre-trains on a large labeled dataset, termed base dataset, and then directly fine-tunes on the target FSL support set, bypassing support-query pairs. Two concurrent works~\cite{Revisiting,Closer} empirically demonstrated that by fixing the backbone network, fine-tuning can achieve competitive performance compared to meta-learning. However, a key challenge lies in the potential inconsistency between the pre-training and downstream FSL tasks, which is especially pronounced when these tasks span highly different domains.
As emphasized in prior works~\cite{metaBaseline, Baseline}, obtaining a high-quality pre-trained model is crucial. To bridge the domain gap and leverage available resources, methods like STARTUP~\cite{selffewshot} not only utilize source data but also incorporate abundant unlabeled target data during training. Li et al.~\cite{li2021universal} address cross-domain challenges by proposing to map domain-specific features into a shared latent space, thereby learning a domain-invariant universal representation.

\subsection{CLIP-based Few-Shot Learning}
Compared to unimodal approaches, multimodal learning in FSL seeks to learn enhanced feature representations by leveraging complementary information across different modalities while minimizing inter-modal redundancies. The emergence of CLIP's robust generalizability has recently spurred significant advances in CLIP-based few-shot learning, primarily through two paradigms: prompt learning/engineering and cache-model learning. For prompt-based methods, CoOp~\cite{CoOp} pioneers few-shot adaptation through learnable prompt vectors that condition CLIP's text encoder on task-specific contexts, while KgCoOp~\cite{KgCoOp} addresses catastrophic forgetting of CLIP's pretrained knowledge by minimizing the euclidean distances between the original handcraft prompts and the learned task-specific prompts. 
% regularizing discrepancies between original and task-adapted prompts via 
% a \textcolor{blue}{knowledge distillation loss}. minimizes the euclidean distance.
Concurrently, ProGard~\cite{ProGrad} selectively updates the prompts whose gradient doesn't conflict with a general direction. 
Though effective, prompts learned in this way often lack interpretability. This motivates prompt engineering methods~\cite{CuLP, CuLP2, CuLP_Hierarchical} that leverage LLMs to generate hierarchical class descriptions. Among them, \cite{CuLP_Hierarchical} computes similarities at each hierarchy level while traversing through the knowledge tree, leading to a summarized score encouraging monotonically increasing confidence. Despite obtaining the hierarchy prompts, these methods directly use the description prompts as classifiers for matching, instead of leveraging them to enhance image representations.

On the other hand, cache-model methods offer an orthogonal direction. 
Tip-adapter~\cite{Tip-adapter} constructs the key-value cache model, while CLIP-Adapter~\cite{Clip-adapter} demonstrates fine-tuning effectiveness with an additional bottleneck layer, that learns new features and performs residual-style feature blending with the original pretrained features.
Subsequent works attempt to 
create mixed-modal prototypes that fuse image cache and the text prompt through self-similar learning and aligned text-visual fusion~\cite{PMPro}. In contrast to the transfer learning discussed in Section~\ref{sec-classical-FSL}, the cache training doesn't necessarily require a base dataset from the same domain, but directly finetunes on the few-shot samples. Building upon these foundations, our approach integrates the cache model with multi-level representation learning, enhancing few-shot classification through cross-modal property alignment. 

% rebuttal里的
\subsection{More Comparative Analysis with Prior Methods}
Our method generates multiple prototypes by averaging the class/property tokens of each class to initialize the cache model. Beyond using a single prototype for each class, only a few methods try to extract multiple prototypes. The method \cite{allen2019infinite} creates multiple prototypes via clustering class supports. Subsequent work~\cite{deuschel2021multi} learns such clusters episodically for histopathology. Meanwhile, \cite{huang2021local} generates multiple prototypes by averaging local expressions with distinct weight matrices from a channel squeeze and spatial excitation (sSE) attention module. Instead of directly weighting the local features, our multiple property tokens are derived from learnable parameters through multi-layer cross-attention modules, with training guided by the selected dominant property descriptions. 
Though Ru et al.~\cite{ru2023token} use a multi-token model, it focuses on alleviating over-smoothing in weakly-supervised segmentation by learning multi-scale semantics without cross-attention. For the attention mechanisms, contemporary vision-language models such as BLIP-2~\cite{li2023blip}, Flamingo~\cite{alayrac2022flamingo} and the OFA~\cite{wang2022ofa} fuse modalities via self-attention with masking, gated cross-attention, or separate positional embeddings. Our MPG module instead uses group-wise FFN after cross-attention between image patches and property queries, complemented by LLM-driven retrieval and contrastive learning.

\section{Preliminaries on CLIP}
\label{sec-prelim}
Before introducing our method, we will briefly review the CLIP framework. The CLIP~\cite{CLIP}, known as a classification foundation model trained on large-scale image-text pairs, consists of an image encoder and a text encoder. The image encoder aims to map high-dimensional images into a low-dimensional embedding space, while the text encoder processes handcrafted text prompts. Through large-scale training, embeddings from the two encoders are aligned with each other, enabling wide applications in classification and image retrieval. 
In particular, given a set of N classes with the template \textit{"A photo of a \{class\}"}, their representations, denoted as $W \in \mathbf{R}^{N\times D}$, are generated via the text encoder. Then for each test image, its classification logits can be formulated as the similarity between the class prompt embeddings $W$ and the test image representation $f_{cls}\in \mathbf{R}^{1\times D}$, that is formulated as below,
\begin{equation}
\label{equ-clip}
    S_{clip}(f_{cls}, W)=f_{cls}W^T \in \mathbf{R}^{1\times N},
\end{equation}
where $f_{cls}$ is the class token extracted by the CLIP's image encoder. In Eq.~\ref{equ-clip}, the CLIP operates only on class names. 

To further enhance the classification, prior-based methods such as Tip-Adapter~\cite{Tip-adapter} build a cache model depending on additional few-shot support images. 
Given the N-class K-shot support training set and the one hot label matrix $F_{label}\in \mathbf{R}^{NK \times N}$, CLIP's image (visual) encoder is used to extract features on support images, denoted as $F_{train}\in \mathbf{R}^{NK \times D}$. The cache model works as an attention module with query, key and value, where visual features of the test image  $f_{cls}$ is the query, the support feature matrix $F_{train}$ is the key, and $F_{label}$ is the value. In this way, the predicted logits of a test image can be obtained as below,  
\begin{equation}
\label{equ-cache}
\begin{split}
S_{cache}(f_{cls},F_{train},F_{label}) =\phi(f_{cls}F_{train}^T)F_{label},
\end{split}
\end{equation}
where $\phi= exp(-\beta \cdot (1-x))$ serves as a non-linear modulator to control the sharpness of the affinity matrix. Here, additional fine-tuning on the key matrix of the cache model, i.e., $F_{train}$, is usually conducted for further performance enhancement. In~\cite{Tip-adapter}, a weighted sum of the cache logits and CLIP's zero-shot scores is calculated for novel class recognition. 

\section{Method}
\label{sec:method}
\subsection{Problem Definition and Overview}

% 缩减版本
Few-shot learning focuses on developing models that can generalize to new tasks with few exemplar images and labels. Given a support set $S=\{(I^s_i,y^s_i)\}_{i=1}^{N\times K}$, where $I^s_i$ is the support image and $y^s_i$ is its label. The goal is to build a model that can accurately predicts the probability distribution $P(y|I^q,S)$, signifying the likelihood of the class label $y$ for a query image $I^q$ given the support set $S$. This is commonly studied in the N-way K-shot setting, where the support set has N classes, each with K examples.

\begin{figure*}[tp]
    \centering
    \includegraphics[width=\linewidth]{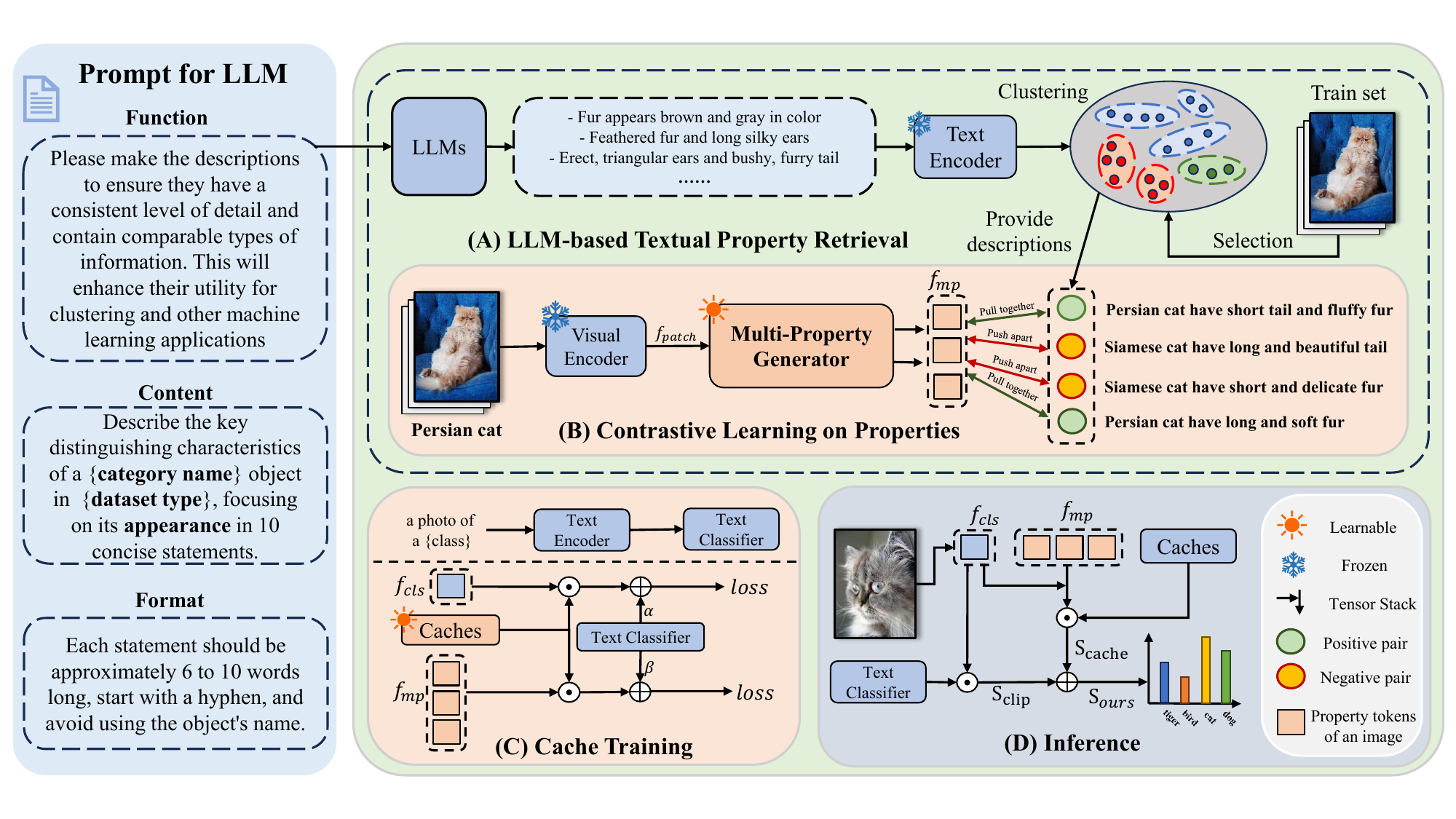}
    \caption{\textbf{The framework of BCT-CLIP}. We introduce a property retrieval process based on LLMs. During training, we employ contrastive learning to generate property tokens, which are aligned with the selected property descriptions. Additionally, we construct caches based on the prototypes that are derived from the training set to enable direct classification.}
    \label{fig:main=framework}
\end{figure*}

In this work, we present \textbf{BCT-CLIP}, a novel method for few-shot learning and classification. We hypothesize that class properties or attributes can provide fine-grained knowledge to enhance classification performance. To this end, we propose generating multiple prototypes to enrich the representation of novel classes using knowledge from Large Language Models (LLMs). The framework is shown in Fig.~\ref{fig:main=framework}.
As mentioned in section~\ref{sec-prelim}, an image can be categorized depending on a cache model derived from the CLIP's class tokens of support images. Upon that, our approach introduces property tokens and new cache models. To this end, \textbf{\textit{this method introduces four new components for training}}: 1) A Multi-Property Generator (MPG) creates new patch-aware tokens for each image, termed \textit{property tokens}. These tokens will subsequently be trained to match class properties; 2) An LLM-based retrieval procedure selects reliable property descriptions as annotations for supervised learning; 3) A new contrastive learning objective aligns property tokens with the above selected fine-grained attributes by pulling the tokens closer to positive property embeddings and pushing them away from negative ones; 4) Class/Property cache models are built and trained to produce the final class prediction. The first three components aim to generate and train property-specific tokens for image representation, while the last is to learn the classification models.
In this way, this framework integrates structural-semantic embeddings, enabling comprehensive image representation and accurate discrimination.

\textbf{\textit{During inference}}, for an unknown image, both its class token and property tokens are achieved via the standard CLIP's visual encoder and our MPG. Note here, our MPG is an efficient plug-in module built upon CLIP. Then the unknown image is predicted to belong to a particular class based on its scores from both the class and property cache models and the zero-shot visual-text similarity. In the following, we will introduce each component and procedure in detail.

\subsection{Multi-Property Generator}\label{sec_mpg}
\begin{figure}[tp]
    \centering
    \includegraphics[width=\linewidth]{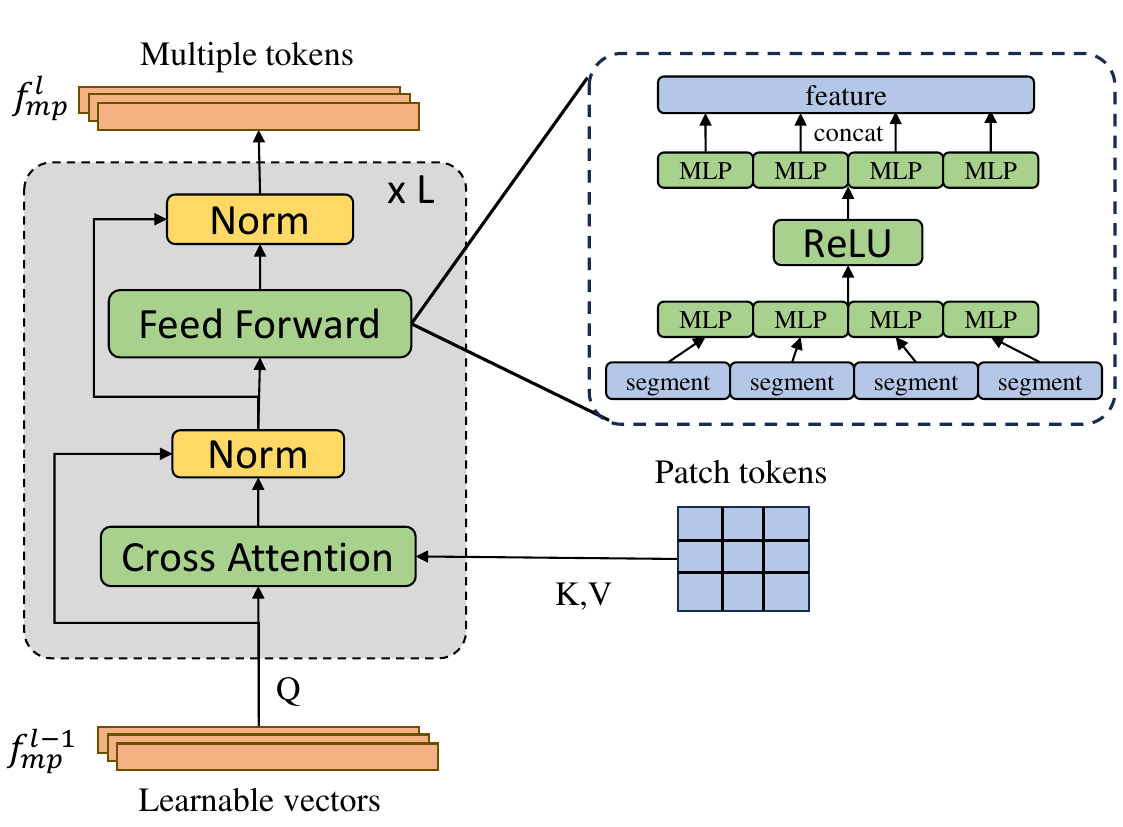}
    \caption{\textbf{The Structure of Multi-property Generator}. It consists of a cross-attention layer, a group-wise Feed Forward Network (FFN), and two norm layers, with residual connections added in between. The input part is made up of patch tokens of the image as K, V, and randomly initialized learnable tokens as Q.}
    \label{fig:MPG}
\end{figure}

We assume that crucial object properties are typically embedded in local regions, such as \textit{beak of birds, fin and scales of fishes}, etc.
Therefore, we propose the Multi-Property Generator (MPG), which generates multiple property-specific visual representations to enrich the embedding space and enables more detailed representations to distinguish subtle visual differences among categories. As shown in Fig.~\ref{fig:MPG}, our MPG is designed as a multi-layer module that processes a set of learnable vectors and image patch features. Specifically, the property-specific tokens are initialized as a set of learnable vectors, termed as property tokens. Through iterative interactions with patch features within the module, these tokens progressively capture task-relevant localized features.

Let's denote $f_{mp} \in \mathbf{R}^{M \times D}$ as the learnable vectors representing property tokens for querying and processing, and $f_{patch} \in \mathbf{R}^{WH \times D}$ the patch features from CLIP's image encoder. Here, $M$ is the query number that is also the number of property tokens, and $W \times H$ is the number of patches. The input of MPG, i.e., the learnable vectors $f_{mp}^0$, are first randomly initialized and then iteratively processed through multiple layers in the module. 
In each layer of MPG, it modulates the query $f_{mp}$ via patch features, in order to generate tokens capturing distinct patch attentions. Specifically, each layer of MPG consists of a cross-attention, normalization layers, and a feed-forward network that is a group-wised two-layer MLP with an activation function\footnote{Instead of standard MLP, group-wised MLP is leveraged for computational efficiency.}. The cross-attention takes $f_{mp}^{l-1}$ from the previous layer as the queries and the same patch features $f_{patch}$ for all layers as the key and value. The process can be formulated as below,
\begin{equation}
    f_{mp}^l=\Phi_l(f_{patch},f_{mp}^{l-1}), l \in [1,L],
\end{equation}
where $\Phi_l$ presents the $l$-th layer computation of the MPG, $L = 2$. The output tokens of each layer, denoted as $f_{mp}^l$, takes the same size as the input vectors $f_{mp}^{l-1}$. In the iterative procedure, $f_{mp}^l$ is further imported into the next layer to serve as the queries and be modulated again. The process continues until all layers are processed. Through the mechanism, the property tokens aggregate information from disparate regions of the images with varying attention. In the subsequent procedure, supervision training based on contrastive learning objectives (Section~\ref{sec:mp-train}) will guild the token learning, in the end align the tokens with semantics of meaningful class attributes. Thanks to the patch-facilitating design, the alignment training becomes feasible even when relying solely on a limited number of support images.

\label{sec:mp-train}

\subsection{Retrieving Textual Property Descriptions from LLM}\label{sec_selection}
To learn meaningful property tokens from MPG, effective supervision is crucial. To this end, we introduce a novel Large Language Model (LLM)-based property retrieval process to achieve a faithful set of class properties, which can be used as ground truth embeddings for supervision. Particularly, our method introduces a novel set of instructions with three key dimensions and a dominant-property selection procedure based on clustering and ranking. 

%缩减
\vspace{3pt}
\subsubsection{Generation of property descriptions by asking LLMs}   
Our approach uses LLMs~\cite{GPT} to generate detailed and meaningful category descriptions, guided by three key instruction dimensions: \textbf{functionality, content, and format}. \textit{An example is shown in Fig.~\ref{fig:main=framework}}. Functionality instructions ensure consistent granularity content for machine learning tasks. This prompt is designed to guide LLMs in generating comparable types of information, thereby facilitating more effective clustering. For content generation, we provide class names and request attribute descriptions that should focus on discriminative appearances but without including the class names. 
Format instructions keep descriptions concise and aligned with a predefined format for batch processing. For example, replacing \{category name\} with ``\textit{tench}" and \{dataset type\} with ``\textit{real world}" generates attributes like ``\textit{Smooth, shiny scale}" and ``\textit{Dorsal fin set far back}". 

LLM-generated text may introduce vague and non-discriminative property descriptions or those that do not match real images, which may undermine the visual representation learning. To avoid this, we proposed a clustering-based procedure for dominant property selection.   

%\vspace{-10px}
\vspace{3pt}
\subsubsection{Semantic clustering of textual descriptions}

Let's encode the descriptions from all categories into a large pool, denoted as $W_{pool}\in \mathbf{R}^{P\times D}$, where $P$ represents the total number of descriptions in the dataset and $D$ is the dimension of text embeddings. With CLIP's textual encoder, we can get features on each property description, and group them into a fixed number of clusters by K-means clustering. The generated cluster set is represented as $W_{cluster}=[W_c^1,W_c^2 \dots W_c^n]$, where $W_c^i \in \mathbf{R}^{P_i \times D}$ denotes the feature set of $i$-th cluster, containing $P_i$ property feature vectors. 
Here, descriptions highlighting similar characteristics are grouped together. 

%\vspace{-10px}
\vspace{3pt}
\subsubsection{Top-M Selection }
Given the property clusters $W_{cluster}$, we leverage CLIP's ability in visual-textual alignment to select property clusters that align with the true content of images within a class. Specifically, for the K-shot support images of a category, we extract their L2-normalized features $f_{clip} \in \mathbf{R}^{K\times D}$ via CLIP's visual encoder. These features are treated as queries to retrieve the top-M nearest property clusters. The cosine similarity between a class and a property cluster  $W_c^i \in \mathbf{R}^{P_i\times D}$ can be calculated as below, 
\begin{equation}\label{eqn-1} 
  S^i_{pcs} =Avg(f_{clip}\cdot (W_c^i)^T),
\end{equation}
where $Avg(\cdot)$ calculates the average value of the similarity matrix between the K support images and $P_i$ property descriptions, and $S^i_{pcs}$ is the similarity score. For each class, we select the top $M$ clusters with the highest scores as the positive property clusters. 
Here note that, the number of selected clusters is kept consistent with the number of property tokens, in order to achieve one-vs-one ground truth embeddings for the M property tokens, thereby enabling supervised learning.  
Specifically, for each class, we will extend the descriptions in the selected clusters with the specific class name for training supervision. Last but not least, the clustering also respects the fact that different categories share similar properties. When combined with ranking, the whole procedure can help identify stable and dominant properties for each class.

\subsection{Contrastive learning on visual property representations}\label{sec_contrast}
Given above achievements, contrastive learning is employed to align the patch-aware tokens generated by MPG, i.e., termed property tokens, with the target property descriptions. This approach effectively mitigates the fine-grained alignment deficiency discussed in Section II and enhances visual representations by integrating semantic knowledge from large language models (LLMs). For each class, we categorize the class-specific description pool into three folds: the positive properties, hard negative properties, and the general negative properties. Our target is to match the image property tokens of a class to their positive descriptions and push them away from all the negative properties. 

\begin{figure}[tp]
    \centering
    \includegraphics[width=\linewidth]{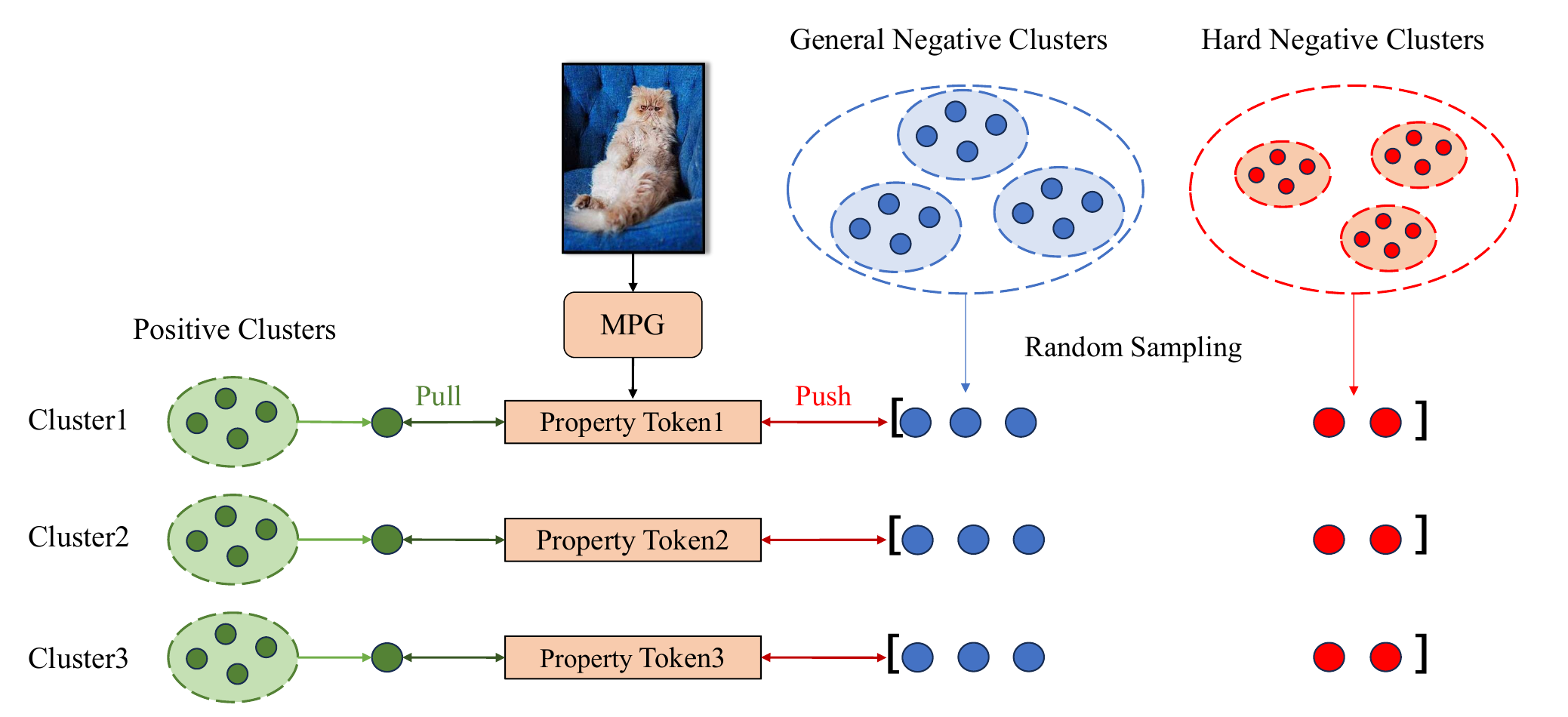}
    \caption{\textbf{The process of selecting positive and general/hard negative properties from clusters. } For each property token, its ground truth positive and negative descriptions are randomly sampled from the corresponding clusters in each training iteration. Here, \textcolor{green}{green}, \textcolor{blue}{blue} and \textcolor{red}{red} dots represent embeddings from positive, general negative, and hard negative property descriptions, respectively.  }
    \label{fig:1v1property}
\end{figure}

In particular, for each class, we define the positive descriptions, hard negatives and general negatives as follows: (1) all the descriptions within a corresponding cluster selected from the Top-M clusters are considered as positives; (2) those positive descriptions from Top-5 confusion categories, that take the most similar characteristics with the target class, are designated as hard negatives; (3) and the positive descriptions from all other classes are viewed as general negatives.   
% 缩减一下
We obtain confusion categories using CLIP's zero-shot scores on training  images, i.e., the limited support images. For effective training, we apply one-vs-one match between each property token and each positive cluster as shown in Fig.~\ref{fig:1v1property}. That is, we assign a numerical label to each token and its corresponding cluster in a sequential manner. Specifically, the first token is consistently mapped to the first cluster, the second token to the second cluster, and so on. By aligning a token with the fixed cluster, we can ensure learning consistent semantics for a specific property token, largely enhanced the training efficacy.

Let's denote the $i$-th property token of the $j$-th image as $f_{mp}^{ij}$.
In each iteration, as shown in Fig.~\ref{fig:1v1property}, we randomly samples one positive property embedding as $W^{ij}_{p} \in \mathbf{R}^{1 \times D}$, $N_{hn}$ hard negative property embeddings as $W^{ij}_{hn} \in \mathbf{R}^{N_{hn} \times D}$, and $N_{gn}$ general negative embeddings as $W^{ij}_{gn} \in \mathbf{R}^{N_{gn} \times D}$, where $N_{hn}$ and $N_{gn}$ are the number of hard negative and general negative properties, respectively. Here we extract features of all the property descriptions via CLIP's textual encoder. The InfoNCE loss applied to the property tokens is formulated as below, 
\begin{equation}
    Loss_{pt}=\sum_{j=1}^{N\times K} \sum_{i=1}^{M} -\log \frac{exp(f^{ij}_{mp}\cdot {W^{ij}_{p}}^T/\tau)}{exp(f^{ij}_{mp}\cdot [W^{ij}_{p},W^{ij}_{hn},W^{ij}_{gn}]^T/\tau)}
\end{equation}
where $[\cdot,\cdot]$ represents concatenation operation,  N, K and M denote category number, shot number per category, and property token number per image, respectively. Our contrastive learning objective is computed by aggregating over all support images across all categories. By randomly sampling positive and negative descriptions from their respective clusters in each training iteration, our contrastive learning maximizes the utilization of the retrieved textual properties.

\subsection{Cache Training}\label{sec_cache}
For direct classification, we construct the cache models based on both the class prototypes and property prototypes, which are respectively achieved by averaging the class and property tokens of the support images from the particular class. Following the format in Eq.~\ref{equ-cache}, we define the class and property cache models as $F_{cache}^{cls}\in \mathbf{R}^{N\times D}$ and  $F_{cache}^{mp} \in \mathbf{R}^{NM\times D}$, respectively, with corresponding label matrices $F_{label}^{cls} \in \mathbf{R}^{N\times N}$ and $F_{label}^{mp} \in \mathbf{R}^{NM\times N} $. Here $N$,$M$ and $D$ are the number of categories, shots per class, and the feature dimension, respectively.  Different from Tip-Adapter\cite{Tip-adapter} that uses all the $N\times K$ support representations, our class cache model use statistical representations of the tokens, i.e., the prototypes, thus taking smaller size.  For practical use, the cache model and label matrix together act as a classifier for the query image. Given a query image $I$, let's denote its class token as $f_{cls}$, its $i$-th property token as $f_{mp}^i$, the classification score on both the cache models are calculated as below, 
\begin{equation}
\begin{split}
    &S_{mp\text{-}cache} = S_{clip}+\alpha \frac{1}{M}\sum_{i=1}^M S_{cache}(f_{mp}^i,F_{cache}^{mp},F_{label}^{mp}),\\
    &S_{cls\text{-}cache} = S_{clip}+\beta S_{cache}(f_{cls},F_{cache}^{cls},F_{label}^{cls}),\\
\end{split}
\label{equ:our-cache-score}
\end{equation}
where $S_{clip}$ is CLIP's zero-shot score vector calculated by Eq.~\ref{equ-clip}, and the function $S_{cache}$ is based on Eq.~\ref{equ-cache}. 
Here,  $\alpha$ and $\beta$ are learnable hyper-parameters that will be optimized in Eq.~\ref{eq-cache-train}. 
As mentioned in Section IV, the cache models are initialized via concatenations of prototype vectors, while the label matrices are concatenations of one-hot label vectors for the prototypes.
For cache finetuning, we leverage the combination of cross-entropy losses on class and property scores for optimization, 
\begin{align}
    Loss_{cache} &= -\sum_{j=1}^{N \times K} y_j [\log(g(S_{cls-cache}))+\nonumber \\ &\log(g(S_{mp-cache}))],
\label{eq-cache-train}
\end{align}
where $g(\cdot)$ is a softmax function, and $y_j$ is the one-hot encoded ground truth label. The cache loss combined with the InfoNCE loss on all property tokens is leveraged for our training. 

\subsection{Inference}
During inference, given a query image $I^q$, we first extract its class token and M property tokens from CLIP and the MPG. By feeding the obtained tokens to the class and property cache models separately, we get the scores on the query image, i.e., $S^q_{mp-cache}$ and $S^q_{cls-cache}$, based on Eq.~\ref{equ:our-cache-score}. We then classify a query image to one of the classes, by adding above two scores. The final inference score $S_{ours}$ is defined as follow,
$$
    S_{ours}=S_{mp-cache}^q+S_{cls-cache}^q.
$$ 
Since $\alpha$ and $\beta$ are both trainable, the score summation does not introduce any hyperparameters.

\section{Experiments and Implementations}
\label{sec:experiment}
In this section, we evaluate our BCT-CLIP across 11 widely recognized benchmark datasets as well as the out-of-distribution data, proving both the effectiveness and robustness of our method. Further ablation studies are also conducted on the ImageNet-1K dataset, demonstrating the contributions of each of our design.

\subsection{Datasets}
Following the previous works~\cite{CoOp,Tip-adapter}, we evaluate the effectiveness of our BCT-CLIP on 11 datasets, including ImageNet\cite{ImageNet}, StandfordCars\cite{StanfordCars}, UCF101\cite{UCF101}, Caltech101\cite{Caltech101}, Flowers102\cite{Flowers102}, SUN397\cite{SUN397}, DTD\cite{DTD}, EuroSAT\cite{EuroSAT}, FGVCAircraft\cite{FGVC}, OxfordPets\cite{OxfordPets} and Food101\cite{Food101}. For the training setting, we exactly follow previous CLIP-based few-shot methods~\cite{CoOp,Tip-adapter}. In particular, we train our Multi-Property Generator (MPG) and the cache models with merely $[1,2,4,8,16]$ training samples, respectively, and test our framework on the full test set.
\begin{figure*}[tp]
    \centering
    \includegraphics[width=\linewidth]{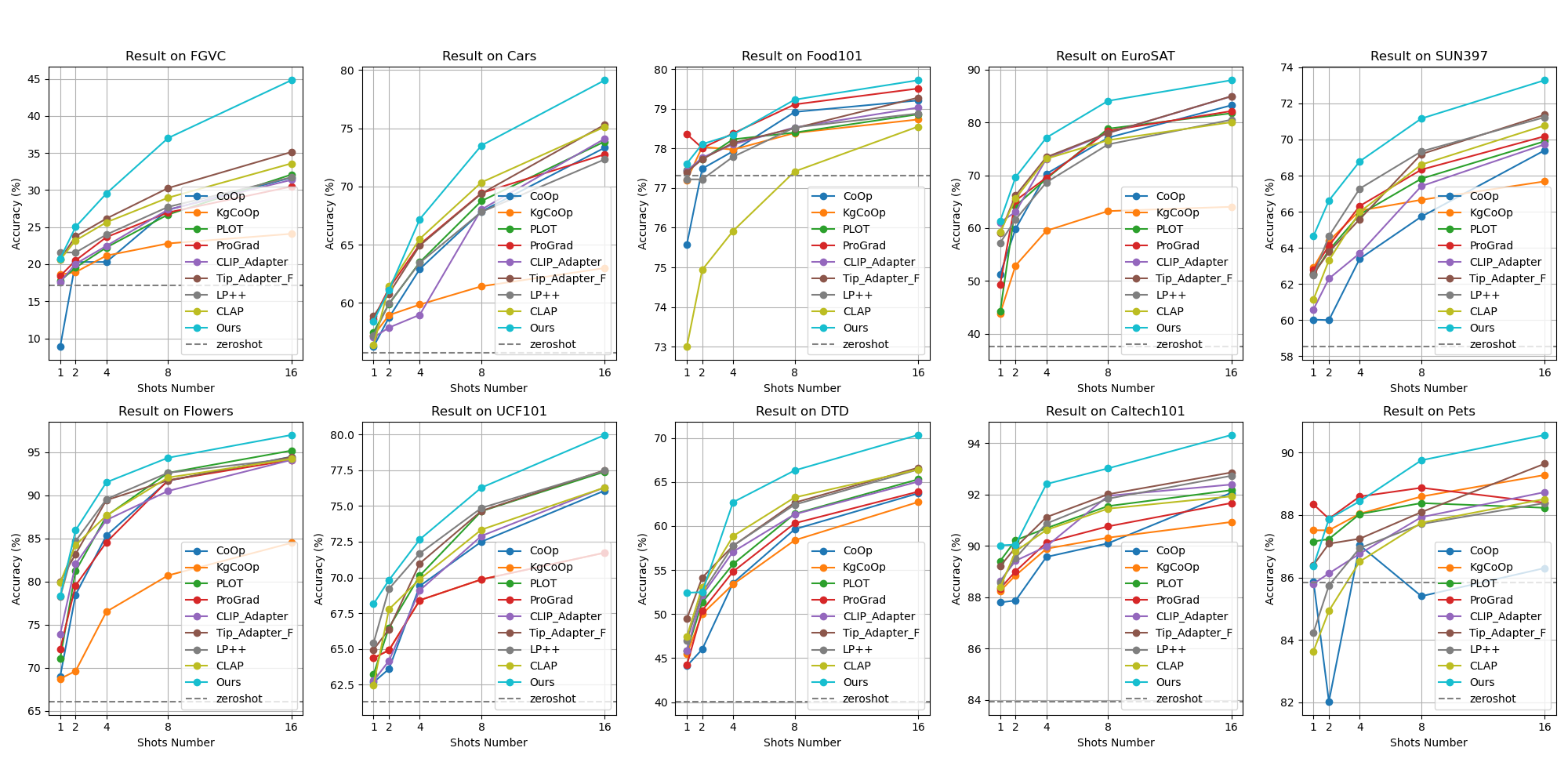}
\caption{Comparison(\%) of different SOTA methods under various few-shot settings on 10 downstream tasks.  Our method outperforms previous methods on most of the downstream tasks, especially when the number of shots increases.}
    \label{fig:10datatset performance}
\end{figure*}

\begin{table}
\begin{center}
\caption{
Comparison (\%) of different SOTA methods on ImageNet-1K under various few-shot settings.
}
\label{tab:ImageNet_Scores}
\begin{tabular}{lccccc}
\hline
Method & 16-shot & 8-shot & 4-shot & 2-shot & 1-shot \\
\hline
CoOp\cite{CoOp} & 62.95 & 61.57 & 59.63 & 57.86 & 57.57 \\ 
KgCoOp\cite{KgCoOp} & 62.43 & 62.20 & 62.00 & 61.44 & 60.90 \\ 
PLOT\cite{PLOT} & 63.08 & 62.48 & 61.79 & 60.73 & 60.46 \\ 
ProGrad\cite{ProGrad} & 63.54 & 63.04 & \underline{62.59} & \underline{62.14} & \underline{61.58} \\ 
CLIP-Adapter\cite{Clip-adapter} & 63.59 & 62.8 & 61.85 & 61.3 & 61.2 \\ 
Tip-Adapter-F\cite{Tip-adapter} & \underline{65.51} & \underline{64.00} & 62.52 & 61.69 & 61.32 \\
LP++\cite{LP++} & 64.73 & 63.76 & 62.55 & 61.56 & 61.18 \\ 
CLAP\cite{CLAP} & 65.02 & 62.98 & 60.73 & 58.50 & 58.50 \\ 
Ours & \textbf{66.40} & \textbf{64.96} & \textbf{63.82} & \textbf{62.42} & \textbf{61.64} \\
\hline
\end{tabular}
\end{center}
\end{table}

\subsection{Implementation Details}
For CLIP, we employ ResNet-50 as the visual encoder and its aligned transformer \cite{attention} as the textual encoder. For zero-shot logits, we input the prompts from multiple templates into the CLIP textual encoder and average the outputs to form the final class textual embedding. Regarding hyperparameters, the number of clusters for k-means clustering of all descriptions is set to half the number of categories in the dataset and the number of property tokens is set to 3, and our MPG has 2 layers. In contrastive learning, the ratio between the positive and negative supervision properties is set to be 1:100, and the ratio between hard and general negatives is initially set as 1:9 and gradually increases to 4:6. In this stage, we train our the MPG for 30 epochs, using the AdamW optimizer with an initial learning rate of 0.0005 and a batch size of 16. 
For cache training, we increase the learning rate to 0.001 via a warm-start strategy and adjust the batch size to 128. The training is performed for 15 epochs with MPG fixed. The entire program is implemented in PyTorch and runs on a single RTX A6000.

\begin{figure}[tp]
    \centering
    \includegraphics[width=\linewidth]{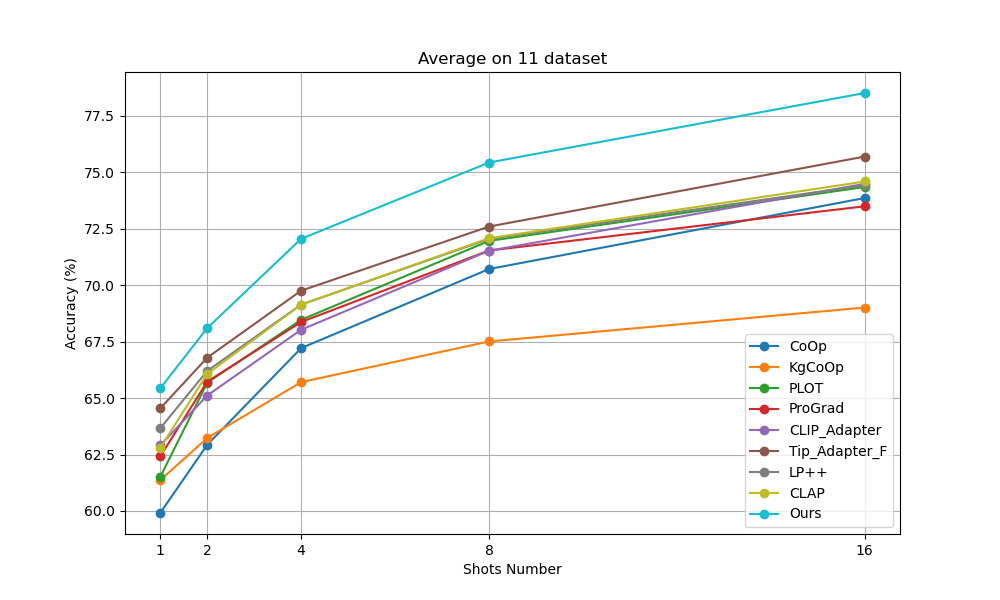}
\caption{Average results on 11 downstream tasks. Our method consistently outperforms previous methods with various number of shots.}
    \label{fig:avg results}
\end{figure}

\section{Results and Analysis}

\subsection{Comparison with State-of-the-Art Methods}
We compare our BCT-CLIP with recent CLIP-based methods on 11 widely recognized datasets.  
For a fair comparison, all these methods are based on the pretrained CLIP with the ResNet-50 visual encoder. 
As shown in Fig.~\ref{fig:10datatset performance} and Tab.~\ref{tab:ImageNet_Scores},
our BCT-CLIP outperforms the baseline methods on most datasets, covering a wide range of scenarios, such as objects, satellite images, textures, and scene images.
Additionally, Fig.~\ref{fig:avg results} illustrates that our model consistently outperforms the others on average across 11 downstream tasks. These results demonstrate the effectiveness and strong generalization capability of our BCT-CLIP.
Our superiority is more obvious when the training shot increases. When we have only one or two shots, we can get slightly better results than Tip-Adapter-F~\cite{Tip-adapter}, which features in finetuning cache models. It shows that even with very limited training data of one shot, our MPG can also provide complementary information for classification. One or two shots, however, may not fully capture the characteristics of a category, leading to limited enhancement. 

\subsection{Out-of-Distribution Performance}
Following previous works \cite{CoOp, Tip-adapter}, we further evaluate the robustness of BCT-CLIP to distribution shifts by training on ``Source" dataset and testing on ``Target" datasets. Specifically, we train models on ImageNet-1K with 16-shot training samples and then test them on out-of-distribution (OOD) benchmarks, that is the ImageNet-V2 \cite{ImageNet-V2} and ImageNet Sketch \cite{ImageNet-Sketch} datasets.
As shown in Tab.~\ref{tab:Cross_Domain}, our model demonstrates strong robustness across these datasets.

\begin{table}
\centering
\caption{Comparison (\%) of different methods under OOD setting. The superior performance of our model demonstrates the robustness and generalizability of our method.}
\label{tab:Cross_Domain}
\setlength{\tabcolsep}{4pt} 
\begin{tabular}{lccc}
\hline
\textbf{Method} & \textbf{ImageNet-1K} & \textbf{ImageNet-V2} & \textbf{ImageNet-Sketch} \\ \hline

CoOp\cite{CoOp} & 62.95 & 55.40 & 34.67 \\ 
PLOT\cite{PLOT} & 63.08 & 55.11 & 33.00 \\
ProGrad\cite{ProGrad} & 63.54 & 54.70 & 34.40 \\ 
CLIP-Adapter\cite{Clip-adapter} & 63.59 & 55.69 & 35.68 \\ 
Tip-Adapter-F\cite{Tip-adapter} & \underline{65.51} & \underline{56.71} & \underline{36.00} \\ 
Ours & \textbf{66.40} & \textbf{57.76} & \textbf{36.54} \\
\hline
\end{tabular}
\end{table}

\begin{table}
\setlength{\tabcolsep}{2pt}
\begin{center}
\caption{Results of BCT-CLIP with different scores combination on ImageNet-1K 16-shot setting.}
\label{tab:ablation_study}
\begin{tabular}{ccccc}
\hline
\textbf{Score} &
\textbf{$S_{clip}$} &
\textbf{$S_{cls-cache}$}& 
\textbf{$S_{mp-cache}$}&
\textbf{$S_{cls-cache}+S_{mp-cache}$}
\\ \hline
\textbf{Accuracy(\%)} & 60.47 & 65.09 & 65.60 & 66.40 \\ \hline

\end{tabular}
\end{center}

\end{table}

\subsection{Ablation Study}

\begin{table*}
\setlength{\tabcolsep}{4pt}
\caption{
Comparison(\%) with zero-shot methods using ResNet50 image encoder.
}
\begin{center}
\begin{tabular}{lccccccccccc}
\hline
Method & ImageNet & StandfordCars & UCF101 & Caltech101 & Flowers102 & SUN397 & DTD & EuroSAT & FGVCAircraft & OxfordPets & Food101 \\
\hline
CLIP\cite{CLIP} & 60.33 & 55.61 & \underline{61.46} & 86.29 & \underline{66.14} & 58.52 & 42.32 & 37.56 & 17.28 & 85.77 & 77.31\\
% \hline
GPT-Hierarchy\cite{CuLP_Hierarchical} & 60.63 & - & - & - & - & - & 45.67 & - & -& 79.99 & 75.44 \\
% \hline
CuLP\cite{CuLP} & \underline{61.67} & \underline{57.28} & 59.71 & \underline{89.17} & 65.97 & \underline{62.75} & \underline{48.46} & \underline{37.99} & \underline{19.53} & \underline{85.04} & \underline{77.07} \\
% \hline
Ours & \textbf{65.34} & \textbf{79.04} & \textbf{79.19} & \textbf{92.86} & \textbf{96.50} & \textbf{71.91} & \textbf{69.20} & \textbf{87.11} & \textbf{42.69} & \textbf{90.65} & \textbf{78.91} \\
\hline
\end{tabular}
\end{center}
\label{tab:zero-shot}
\end{table*}

\paragraph{Effectiveness of Various Components} To evaluate the effect of different modules on BCT-CLIP, we performed ablation studies on the ImageNet-1K dataset. Specifically, Tab.~\ref{tab:ablation_study} presents the results of BCT-CLIP with various components. Compared to results gained on CLIP zero-shot score of $S_{clip}$, the performance based on $S_{mp-cache}$, a combination of $S_{clip}$ and scores based on property tokens, gets a significant improvement, with accuracy increase of 5.13\%. Additionally, the score based on property tokens also outperforms that based on class tokens. And it demonstrates that incorporating a multi-property cache alone is sufficient to surpass the class-cache-based benchmark~\cite{Tip-adapter}. 
Furthermore, by adding property-cache score upon the class-cache score, our model can get additional performance improvement. This enhancement is observed both before and after fine-tuning the property cache, which clearly demonstrates that our approach can successfully learn complementary knowledge that supplements the class tokens.

\paragraph{Effectiveness of Various Backbones}
More comparisons on \textbf{various image encoders} are shown in Tab.~\ref{tab:backbone}, BCT-CLIP consistently outperforms the baseline methods.

\begin{table}
\caption{
Comparison(\%) with SOTA methods on ImageNet-1K with 16-shot setting.
}
\begin{center}
\begin{tabular}{lcccc}
\hline
Method & ResNet50 & ResNet101 & ViT-B/32 & ViT-B/16 \\
\hline
CLIP & 60.33 & 62.53 & 63.80 & 68.73\\ 
% \hline
CLIP-Adapter & 63.59 & 65.39 & 66.19 & 71.13 \\
% \hline
Tip-Adapter-F & \underline{65.51} & \underline{68.56} & \underline{68.65} & \underline{73.69} \\
% \hline
Ours & \textbf{66.45} & \textbf{69.28} & \textbf{69.88} & \textbf{75.08} \\
\hline

\end{tabular}

\label{tab:backbone}
\end{center}
\end{table}

\paragraph{Robustness on Hyperparameters} Regarding hyperparameters in contrastive learning, we use a stable contrastive sampling strategy as described in Sec.~\ref{sec_contrast}, which has been proven consistently effective across a range of datasets. This could be due to the distinct split of \{hard, general\}-negatives compared to the positive properties. 

For the InfoNCE loss, the temperature parameter is 0.3. In addition, we studied on the number of property tokens using a validation set of 5K images from the training dataset, and determined property number M=3. For further validation, we evaluated models with 1, 3, and 5 property tokens, achieving accuracies of 66.20\%, 66.40\%, and 66.30\%, respectively, which demonstrates that incorporating additional property tokens can enhance performance.

Note that, since the $\alpha$ and $\beta$ in Eq.~\ref{equ:our-cache-score} are learnable, which do not introduce additional hyperparameters. As shown in Tab.~\ref{tab:hyper}, the final learned values of \textbf{$\alpha$} and \textbf{$\beta$} vary across datasets but with more weights on property tokens, revealing the importance of the LLM-derived local property.

\begin{table}
\caption{
The values of $\alpha$ and $\beta$ on different datasets after training. Larger $\alpha$ indicates more weights on property tokens. 
}
\begin{center}
\begin{tabular}{lcccc}
\hline
dataset & ImageNet & FGVC & Caltech101 & StanfordCars \\\hline
$\alpha$ / $\beta$ & 0.39/0.09 & 0.65/0.62 & 0.5/0.27 & 0.41/0.20 \\\hline
\end{tabular}
\label{tab:hyper}
\end{center}
\end{table}

\paragraph{Analysis of Complexity} 

Unlike previous methods using all training data to build the cache model, ours uses only average class and property tokens per class, reducing parameters and computation. MPG, a small module with two layer attention and group-wise FFN, adds just 0.92 GFLOPS at test time. With prepared data, MPG takes about 8.2 minutes to update via the 30-epoch training.

\begin{figure}[tp]
    \centering
    \includegraphics[width=\linewidth]{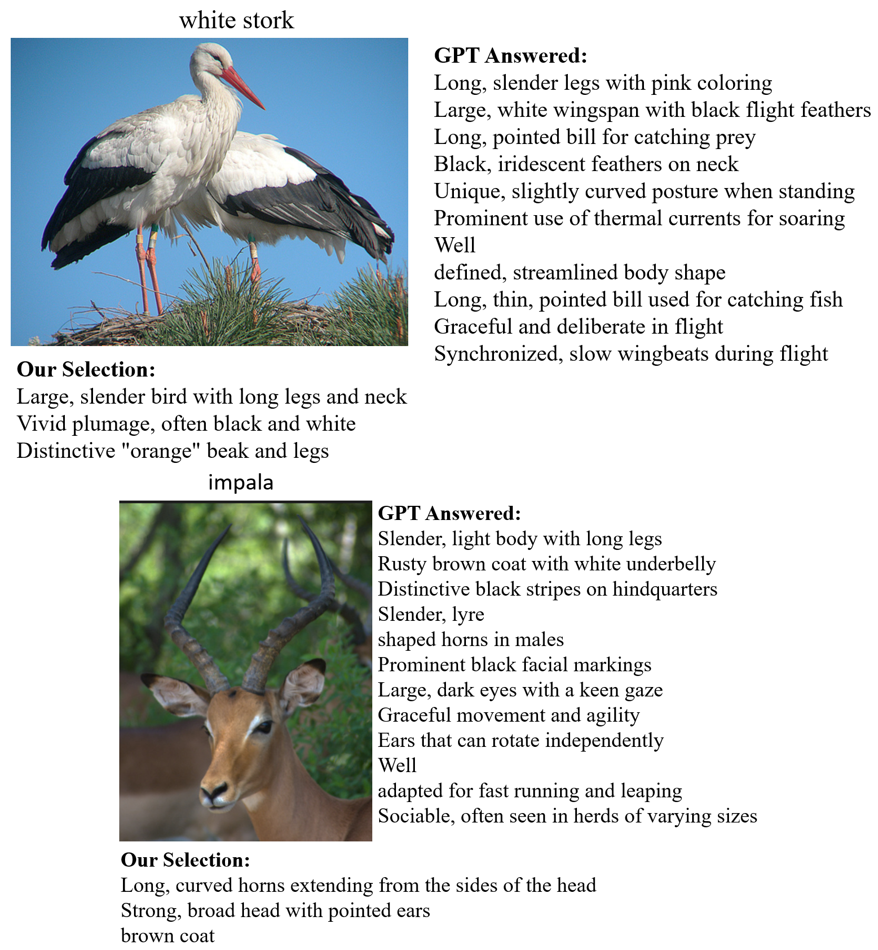}
\caption{Comparison between our selected descriptions and those generated directly using GPT 3.5. Our selection identifies dominant properties that are unique and specifically tied to visual appearance.}
    \label{fig:property-visulize}
\end{figure}

\paragraph{Analysis on Property Description Selection} As shown in Fig.~\ref{fig:property-visulize}, our property selection process demonstrates distinct advantages by leveraging clustering and image similarity to select properties grounded in context and visual content. 
While LLM-generated descriptions based on class names  mainly focus on general traits (e.g., the ``long, thin, pointed bill"  that is common among diverse bird species), our method excels at identifying unique and specific features tied directly to the visual content, as shown in Fig.~\ref{fig:property-visulize}. 
Our descriptions capture details traits like ``orange'' beak and legs of white storks in the upper image and the ``long, curved'' horns of impalas in the bottom image. In addition, our selection can also filter out descriptions that are too challenging to be captured or defined in appearance, such as ``prominent use of thermal currents for soaring'' and ``graceful and deliberate in flight''. 
By retaining common features of the object class while highlighting distinctive traits, our approach provides precise and meaningful attributes that adapt effectively to diverse visual contexts. 

% \vspace{-10pt}
\paragraph{Analysis on Property Tokens} 

To better illustrate the property tokens, we calculate the angular distances among the $M$ property tokens across the training images.
When $M$ is set to 3, the average mutual angular distances among learned property tokens are $12.0^\circ$, $12.89^\circ$, and $9.6^\circ$, respectively, indicating noticeable differences in the attribute features extracted from the images.

In addition, we analyze the necessity of learning visual property representations by comparing our method with the methods based on only prompt engineering~\cite{CuLP, CuLP_Hierarchical}. The baseline methods directly leverage CLIP's embeddings on descriptive sentences~\cite{CuLP} and the descriptions of class hierarchy~\cite{CuLP_Hierarchical} to classify a global image representation. The comparison is shown in Tab.~\ref{tab:zero-shot}. For a fair comparison, we report the results based on the initialized cache models which are not trained. 
These methods typically classify an image in a zero-shot manner by comparing the text embeddings of their refined detailed descriptions with the image class token. 
Our results in Tab.~\ref{tab:zero-shot} aim to demonstrate that \textit{direct text-image matching based on class tokens is not enough}, despite the text descriptions being enhanced to include more detailed concepts. Training visual property tokens is still necessary, which can lead to more precise matching on the same conceptual levels.

\section{Conclusion}
\label{sec::conclusion}
In summary, we present BCT-CLIP, an innovative few-shot learning approach that enhances image representations by leveraging additional knowledge from Large-Language Models (LLMs). Specifically, for the new representations, we introduce a novel Multi-Property Generator (MPG) to create property tokens, an LLM-based property description selection procedure to achieve training objectives, and a new contrastive learning procedure to align the generated property tokens with target property descriptions. By aligning the trained property tokens with fine-grained semantics, our method effectively addresses the \textit{fine-grained alignment deficiency} issue that occurs for CLIP's class tokens. Building on both class and property tokens, we further train cache models to achieve accurate classification. Extensive experiments conducted on 11 widely used datasets demonstrate the efficacy of our proposed method. Additional out-of-distribution performance evaluations attest to the robustness and generalizability of the learned model.

\bibliographystyle{IEEEtran}
\bibliography{IEEEabrv,./paper}

\begin{thebibliography}{10}
\providecommand{\url}[1]{#1}
\csname url@samestyle\endcsname
\providecommand{\newblock}{\relax}
\providecommand{\bibinfo}[2]{#2}
\providecommand{\BIBentrySTDinterwordspacing}{\spaceskip=0pt\relax}
\providecommand{\BIBentryALTinterwordstretchfactor}{4}
\providecommand{\BIBentryALTinterwordspacing}{\spaceskip=\fontdimen2\font plus
\BIBentryALTinterwordstretchfactor\fontdimen3\font minus \fontdimen4\font\relax}
\providecommand{\BIBforeignlanguage}[2]{{%
\expandafter\ifx\csname l@#1\endcsname\relax
\typeout{** WARNING: IEEEtran.bst: No hyphenation pattern has been}%
\typeout{** loaded for the language `#1'. Using the pattern for}%
\typeout{** the default language instead.}%
\else
\language=\csname l@#1\endcsname
\fi
#2}}
\providecommand{\BIBdecl}{\relax}
\BIBdecl

\bibitem{koch2015siamese}
G.~Koch, R.~Zemel, R.~Salakhutdinov \emph{et~al.}, ``Siamese neural networks for one-shot image recognition,'' in \emph{ICML deep learning workshop}, vol.~2, no.~1.\hskip 1em plus 0.5em minus 0.4em\relax Lille, 2015, pp. 1--30.

\bibitem{matching_net}
O.~Vinyals, C.~Blundell, T.~Lillicrap, D.~Wierstra \emph{et~al.}, ``Matching networks for one shot learning,'' pp. 3630--3638, 2016.

\bibitem{protonet}
J.~Snell, K.~Swersky, and R.~Zemel, ``Prototypical networks for few-shot learning,'' \emph{Advances in neural information processing systems}, vol.~30, 2017.

\bibitem{u_net}
O.~Ronneberger, P.~Fischer, and T.~Brox, ``U-net: Convolutional networks for biomedical image segmentation,'' in \emph{Medical Image Computing and Computer-Assisted Intervention--MICCAI 2015: 18th International Conference, Munich, Germany, October 5-9, 2015, Proceedings, Part III 18}.\hskip 1em plus 0.5em minus 0.4em\relax Springer, 2015, pp. 234--241.

\bibitem{meta_rcnn}
X.~Yan, Z.~Chen, A.~Xu, X.~Wang, X.~Liang, and L.~Lin, ``Meta r-cnn: Towards general solver for instance-level low-shot learning,'' in \emph{Proceedings of the IEEE/CVF International Conference on Computer Vision}, 2019, pp. 9577--9586.

\bibitem{fsod}
Q.~Fan, W.~Zhuo, C.-K. Tang, and Y.-W. Tai, ``Few-shot object detection with attention-rpn and multi-relation detector,'' in \emph{Proceedings of the IEEE/CVF conference on computer vision and pattern recognition}, 2020, pp. 4013--4022.

\bibitem{latent-mining}
L.~Yang, W.~Zhuo, L.~Qi, Y.~Shi, and Y.~Gao, ``Mining latent classes for few-shot segmentation,'' in \emph{Proceedings of the IEEE/CVF international conference on computer vision}, 2021, pp. 8721--8730.

\bibitem{alexnet}
A.~Krizhevsky, I.~Sutskever, and G.~E. Hinton, ``Imagenet classification with deep convolutional neural networks,'' \emph{Advances in neural information processing systems}, vol.~25, 2012.

\bibitem{vgg}
K.~Simonyan and A.~Zisserman, ``Very deep convolutional networks for large-scale image recognition,'' \emph{arXiv preprint arXiv:1409.1556}, 2014.

\bibitem{ResNet}
K.~He, X.~Zhang, S.~Ren, and J.~Sun, ``Deep residual learning for image recognition,'' in \emph{Proceedings of the IEEE conference on computer vision and pattern recognition}, 2016, pp. 770--778.

\bibitem{CoOp}
K.~Zhou, J.~Yang, C.~C. Loy, and Z.~Liu, ``Learning to prompt for vision-language models,'' \emph{International Journal of Computer Vision}, vol. 130, no.~9, pp. 2337--2348, 2022.

\bibitem{zhou2022conditional}
K.~Zhou, J.~Yang, C.~Loy, and Z.~Liu, ``Conditional prompt learning for vision-language models,'' in \emph{Proceedings of the IEEE/CVF conference on computer vision and pattern recognition}, 2022, pp. 16\,816--16\,825.

\bibitem{CuLP}
S.~Pratt, I.~Covert, R.~Liu, and A.~Farhadi, ``What does a platypus look like? generating customized prompts for zero-shot image classification,'' in \emph{Proceedings of the IEEE/CVF International Conference on Computer Vision}, 2023, pp. 15\,691--15\,701.

\bibitem{CuLP_Hierarchical}
Z.~Ren, Y.~Su, and X.~Liu, ``Chatgpt-powered hierarchical comparisons for image classification,'' \emph{Advances in Neural Information Processing Systems}, vol.~36, 2024.

\bibitem{Tip-adapter}
R.~Zhang, R.~Fang, W.~Zhang, P.~Gao, K.~Li, J.~Dai, Y.~Qiao, and H.~Li, ``Tip-adapter: Training-free clip-adapter for better vision-language modeling,'' \emph{arXiv preprint arXiv:2111.03930}, 2021.

\bibitem{tipx}
V.~Udandarao, A.~Gupta, and S.~Albanie, ``Sus-x: Training-free name-only transfer of vision-language models,'' in \emph{Proceedings of the IEEE/CVF International Conference on Computer Vision}, 2023, pp. 2725--2736.

\bibitem{PMPro}
Y.~Su, X.~Liu, Y.~Zhao, R.~Hong, and M.~Wang, ``Partial-tuning based mixed-modal prototypes for few-shot classification,'' \emph{IEEE Transactions on Multimedia}, vol.~26, pp. 9175--9186, 2024.

\bibitem{munkhdalai2017meta}
T.~Munkhdalai and H.~Yu, ``Meta networks,'' in \emph{Proceedings of the 34th International Conference on Machine Learning-Volume 70}.\hskip 1em plus 0.5em minus 0.4em\relax JMLR. org, 2017, pp. 2554--2563.

\bibitem{rombach2022high}
R.~Rombach, A.~Blattmann, D.~Lorenz, P.~Esser, and B.~Ommer, ``High-resolution image synthesis with latent diffusion models,'' in \emph{Proceedings of the IEEE/CVF conference on computer vision and pattern recognition}, 2022, pp. 10\,684--10\,695.

\bibitem{Reptile}
A.~Nichol, J.~Achiam, and J.~Schulman, ``On first-order meta-learning algorithms,'' \emph{arXiv preprint arXiv:1803.02999}, 2018.

\bibitem{zhu2020attribute}
Y.~Zhu, W.~Min, and S.~Jiang, ``Attribute-guided feature learning for few-shot image recognition,'' \emph{IEEE Transactions on Multimedia}, vol.~23, pp. 1200--1209, 2020.

\bibitem{Revisiting}
A.~Nakamura and T.~Harada, ``Revisiting fine-tuning for few-shot learning,'' \emph{arXiv preprint arXiv:1910.00216}, 2019.

\bibitem{Closer}
W.-Y. Chen, Y.-C. Liu, Z.~Kira, Y.-C. Wang, and J.-B. Huang, ``A closer look at few-shot classification,'' in \emph{International Conference on Learning Representations}, 2019.

\bibitem{metaBaseline}
Y.~Chen, Z.~Liu, H.~Xu, T.~Darrell, and X.~Wang, ``Meta-baseline: Exploring simple meta-learning for few-shot learning,'' in \emph{Proceedings of the IEEE/CVF international conference on computer vision}, 2021, pp. 9062--9071.

\bibitem{Baseline}
G.~S. Dhillon, P.~Chaudhari, A.~Ravichandran, and S.~Soatto, ``A baseline for few-shot image classification,'' 2019.

\bibitem{selffewshot}
C.~P. Phoo and B.~Hariharan, ``Self-training for few-shot transfer across extreme task differences,'' \emph{arXiv:2010.07734}, 2020.

\bibitem{li2021universal}
W.-H. Li, X.~Liu, and H.~Bilen, ``Universal representation learning from multiple domains for few-shot classification,'' in \emph{Proceedings of the IEEE/CVF international conference on computer vision}, 2021, pp. 9526--9535.

\bibitem{KgCoOp}
H.~Yao, R.~Zhang, and C.~Xu, ``Visual-language prompt tuning with knowledge-guided context optimization,'' in \emph{Proceedings of the IEEE/CVF conference on computer vision and pattern recognition}, 2023, pp. 6757--6767.

\bibitem{ProGrad}
B.~Zhu, Y.~Niu, Y.~Han, Y.~Wu, and H.~Zhang, ``Prompt-aligned gradient for prompt tuning,'' in \emph{Proceedings of the IEEE/CVF international conference on computer vision}, 2023, pp. 15\,659--15\,669.

\bibitem{CuLP2}
S.~Menon and C.~Vondrick, ``Visual classification via description from large language models,'' 2023.

\bibitem{Clip-adapter}
P.~Gao, S.~Geng, R.~Zhang, T.~Ma, R.~Fang, Y.~Zhang, H.~Li, and Y.~Qiao, ``Clip-adapter: Better vision-language models with feature adapters,'' \emph{International Journal of Computer Vision}, vol. 132, no.~2, pp. 581--595, 2024.

\bibitem{allen2019infinite}
K.~Allen, E.~Shelhamer, H.~Shin, and J.~Tenenbaum, ``Infinite mixture prototypes for few-shot learning,'' in \emph{International conference on machine learning}.\hskip 1em plus 0.5em minus 0.4em\relax PMLR, 2019, pp. 232--241.

\bibitem{deuschel2021multi}
J.~Deuschel, D.~Firmbach, C.~I. Geppert, M.~Eckstein, A.~Hartmann, V.~Bruns, P.~Kuritcyn, J.~Dexl, D.~Hartmann, D.~Perrin \emph{et~al.}, ``Multi-prototype few-shot learning in histopathology,'' in \emph{Proceedings of the IEEE/CVF international conference on computer vision}, 2021, pp. 620--628.

\bibitem{huang2021local}
H.~Huang, Z.~Wu, W.~Li, J.~Huo, and Y.~Gao, ``Local descriptor-based multi-prototype network for few-shot learning,'' \emph{Pattern Recognition}, vol. 116, p. 107935, 2021.

\bibitem{ru2023token}
L.~Ru, H.~Zheng, Y.~Zhan, and B.~Du, ``Token contrast for weakly-supervised semantic segmentation,'' in \emph{Proceedings of the IEEE/CVF Conference on Computer Vision and Pattern Recognition}, 2023, pp. 3093--3102.

\bibitem{li2023blip}
J.~Li, D.~Li, S.~Savarese, and S.~Hoi, ``Blip-2: Bootstrapping language-image pre-training with frozen image encoders and large language models,'' in \emph{International conference on machine learning}.\hskip 1em plus 0.5em minus 0.4em\relax PMLR, 2023, pp. 19\,730--19\,742.

\bibitem{alayrac2022flamingo}
J.-B. Alayrac, J.~Donahue, P.~Luc, A.~Miech, I.~Barr, Y.~Hasson, K.~Lenc, A.~Mensch, K.~Millican, M.~Reynolds \emph{et~al.}, ``Flamingo: a visual language model for few-shot learning,'' \emph{Advances in neural information processing systems}, vol.~35, pp. 23\,716--23\,736, 2022.

\bibitem{wang2022ofa}
P.~Wang, A.~Yang, R.~Men, J.~Lin, S.~Bai, Z.~Li, J.~Ma, C.~Zhou, J.~Zhou, and H.~Yang, ``Ofa: Unifying architectures, tasks, and modalities through a simple sequence-to-sequence learning framework,'' in \emph{International conference on machine learning}.\hskip 1em plus 0.5em minus 0.4em\relax PMLR, 2022, pp. 23\,318--23\,340.

\bibitem{CLIP}
A.~Radford, J.~W. Kim, C.~Hallacy, A.~Ramesh, G.~Goh, S.~Agarwal, G.~Sastry, A.~Askell, P.~Mishkin, J.~Clark \emph{et~al.}, ``Learning transferable visual models from natural language supervision,'' in \emph{International conference on machine learning}.\hskip 1em plus 0.5em minus 0.4em\relax PMLR, 2021, pp. 8748--8763.

\bibitem{GPT}
T.~Brown, B.~Mann, N.~Ryder, M.~Subbiah, J.~D. Kaplan, P.~Dhariwal, A.~Neelakantan, P.~Shyam, G.~Sastry, A.~Askell \emph{et~al.}, ``Language models are few-shot learners,'' \emph{Advances in neural information processing systems}, vol.~33, pp. 1877--1901, 2020.

\bibitem{ImageNet}
J.~Deng, W.~Dong, R.~Socher, L.-J. Li, K.~Li, and L.~Fei-Fei, ``Imagenet: A large-scale hierarchical image database,'' in \emph{2009 IEEE Conference on Computer Vision and Pattern Recognition}, 2009, pp. 248--255.

\bibitem{StanfordCars}
J.~Krause, M.~Stark, J.~Deng, and L.~Fei-Fei, ``3d object representations for fine-grained categorization,'' in \emph{2013 IEEE International Conference on Computer Vision Workshops}, 2013, pp. 554--561.

\bibitem{UCF101}
\BIBentryALTinterwordspacing
K.~Soomro, A.~R. Zamir, and M.~Shah, ``Ucf101: A dataset of 101 human actions classes from videos in the wild,'' 2012. [Online]. Available: \url{https://arxiv.org/abs/1212.0402}
\BIBentrySTDinterwordspacing

\bibitem{Caltech101}
L.~Fei-Fei, R.~Fergus, and P.~Perona, ``Learning generative visual models from few training examples: An incremental bayesian approach tested on 101 object categories,'' in \emph{2004 Conference on Computer Vision and Pattern Recognition Workshop}, 2004, pp. 178--178.

\bibitem{Flowers102}
M.-E. Nilsback and A.~Zisserman, ``Automated flower classification over a large number of classes,'' in \emph{2008 Sixth Indian conference on computer vision, graphics \& image processing}.\hskip 1em plus 0.5em minus 0.4em\relax IEEE, 2008, pp. 722--729.

\bibitem{SUN397}
J.~Xiao, J.~Hays, K.~A. Ehinger, A.~Oliva, and A.~Torralba, ``Sun database: Large-scale scene recognition from abbey to zoo,'' in \emph{2010 IEEE Computer Society Conference on Computer Vision and Pattern Recognition}, 2010, pp. 3485--3492.

\bibitem{DTD}
M.~Cimpoi, S.~Maji, I.~Kokkinos, S.~Mohamed, and A.~Vedaldi, ``Describing textures in the wild,'' in \emph{Proceedings of the IEEE conference on computer vision and pattern recognition}, 2014, pp. 3606--3613.

\bibitem{EuroSAT}
P.~Helber, B.~Bischke, A.~Dengel, and D.~Borth, ``Eurosat: A novel dataset and deep learning benchmark for land use and land cover classification,'' \emph{IEEE Journal of Selected Topics in Applied Earth Observations and Remote Sensing}, vol.~12, no.~7, pp. 2217--2226, 2019.

\bibitem{FGVC}
\BIBentryALTinterwordspacing
S.~Maji, E.~Rahtu, J.~Kannala, M.~Blaschko, and A.~Vedaldi, ``Fine-grained visual classification of aircraft,'' 2013. [Online]. Available: \url{https://arxiv.org/abs/1306.5151}
\BIBentrySTDinterwordspacing

\bibitem{OxfordPets}
O.~M. Parkhi, A.~Vedaldi, A.~Zisserman, and C.~V. Jawahar, ``Cats and dogs,'' in \emph{2012 IEEE Conference on Computer Vision and Pattern Recognition}, 2012, pp. 3498--3505.

\bibitem{Food101}
L.~Bossard, M.~Guillaumin, and L.~Van~Gool, ``Food-101 -- mining discriminative components with random forests,'' in \emph{Computer Vision -- ECCV 2014}, D.~Fleet, T.~Pajdla, B.~Schiele, and T.~Tuytelaars, Eds.\hskip 1em plus 0.5em minus 0.4em\relax Cham: Springer International Publishing, 2014, pp. 446--461.

\bibitem{PLOT}
G.~Chen, W.~Yao, X.~Song, X.~Li, Y.~Rao, and K.~Zhang, ``Plot: Prompt learning with optimal transport for vision-language models,'' in \emph{ICLR}, 2023.

\bibitem{LP++}
Y.~Huang, F.~Shakeri, J.~Dolz, M.~Boudiaf, H.~Bahig, and I.~Ben~Ayed, ``Lp++: A surprisingly strong linear probe for few-shot clip,'' in \emph{Proceedings of the IEEE/CVF Conference on Computer Vision and Pattern Recognition}, 2024, pp. 23\,773--23\,782.

\bibitem{CLAP}
J.~Silva-Rodriguez, S.~Hajimiri, I.~Ben~Ayed, and J.~Dolz, ``A closer look at the few-shot adaptation of large vision-language models,'' in \emph{Proceedings of the IEEE/CVF Conference on Computer Vision and Pattern Recognition}, 2024, pp. 23\,681--23\,690.

\bibitem{attention}
A.~Vaswani, N.~Shazeer, N.~Parmar, J.~Uszkoreit, L.~Jones, A.~N. Gomez, {\L}.~Kaiser, and I.~Polosukhin, ``Attention is all you need,'' \emph{Advances in neural information processing systems}, vol.~30, 2017.

\bibitem{ImageNet-V2}
B.~Recht, R.~Roelofs, L.~Schmidt, and V.~Shankar, ``Do imagenet classifiers generalize to imagenet?'' in \emph{International conference on machine learning}.\hskip 1em plus 0.5em minus 0.4em\relax PMLR, 2019, pp. 5389--5400.

\bibitem{ImageNet-Sketch}
H.~Wang, S.~Ge, Z.~Lipton, and E.~P. Xing, ``Learning robust global representations by penalizing local predictive power,'' \emph{Advances in neural information processing systems}, vol.~32, 2019.

\end{thebibliography}

\vfill

\end{document}